# Stochastic Complexity of Bayesian Networks

Keisuke Yamazaki
Department of Computational Intelligence
and Systems Science,
Tokyo Institute of Technology,
4259 Nagatsuta, Midori-ku, Yokohama,
226-8503 Japan

Sumio Watanabe
Precision and Intelligence Laboratory,
Tokyo Institute of Technology,
4259 Nagatsuta, Midori-ku, Yokohama,
226-8503 Japan

## Abstract

Bayesian networks are now used in enormous fields, for example, system diagnosis, data mining, clusterings etc. In spite of wide range of their applications, the statistical properties have not yet been clarified because the models are nonidentifiable and non-regular. In a Bayesian network, the set of parameters for a smaller model is an analytic set with singularities in the parameter space of a large model. Because of these singularities, the Fisher information matrices are not positive definite. In other words, the mathematical foundation for learning has not been constructed. In recent years, however, we have developed a method to analyze non-regular models by using algebraic geometry. This method revealed the relation between model's singularities and its statistical properties. In this paper, applying this method to Bayesian networks with latent variables, we clarify the orders of the stochastic complexities. Our result shows that their upper bound is smaller than the dimension of the parameter space. This means that the Bayesian generalization error is also far smaller than that of a regular model, and that Schwarz's model selection criterion BIC needs to be improved for Bayesian networks.

## 1 Introduction

Recently, Bayesian networks have been widely used in information processing and uncertain artificial intelligence. For example, Bayesian networks are applied to data mining, system fault diagnosis, and software accessibility options. In spite of these applications and many training algorithms, their statistical properties such as the generalization error have not yet been clarified.

All learning models fall into two types. One is *identifiable*, the other is *non-identifiable*. In general, the learning model is described by the probability density function $p(x|w)$, where $w$ is the parameter. If the mapping from the parameter to the probability density function is one-to-one, then the model is called *identifiable*, otherwise, *non-identifiable*.

One of the difficulties in the analysis of the non-identifiable model is that we cannot apply the method of a regular model to a non-identifiable one. If the learning model attains the true distribution from which sample data are taken, the true parameter is not one point but an analytic set in the parameter space. The set generally includes many singularities. Because of these singularities, the Fisher information matrices are not positive definite. This means that the log likelihood cannot be approximated by any quadratic form of the parameter in the neighborhood of these singularities. This is why the mathematical properties of the non-identifiable models have been unknown.

Bayesian networks are non-identifiable models as are many models used information engineering fields, such as multi-layered perceptrons, mixture models, and Boltzmann machines. Let us illustrate the singularities by the simplest example. Assume that the true distribution is defined by

$$q(x) = b^{*x} \times (1 - b^*)^{1-x},$$

where $0 \leq b^* \leq 1$, and we define $0^0 = 1$. This distribution has one observable node $x \in \{0, 1\}$ and no latent nodes. Also assume that a learning machine is defined by

$$\begin{aligned}
p(x|a, b_1, b_2) &= p(h=1)p(x|h=1) \\
&\quad + p(h=0)p(x|h=0) \\
&= a\left(b_1^x \times (1-b_1)^{1-x}\right) \\
&\quad + (1-a)\left(b_2^x \times (1-b_2)^{1-x}\right),
\end{aligned}$$

where $0 \leq a, b_1, b_2 \leq 1$. This model has one observable node $x$ and one latent node $h \in \{0,1\}$ (Figure 1, (a)). In this model, the set of the true parameters is

$$\{a = 1, b_1 = b^*\} \cup \{a = 0, b_2 = b^*\} \cup \{b_1 = b_2 = b^*\}.$$

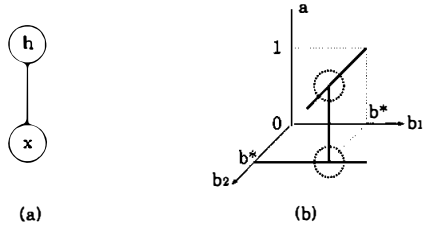

Figure 1: The simplest example. (a) The learning model has one observable node and one hidden node. (b) The singularities in the parameter space.

This set has singularities (Figure 1, (b)),

$$(a, b_1, b_2) = (1, b^*, b^*), (0, b^*, b^*).$$

Though this is the simplest case, we cannot analyze it by the regular model method.

The importance of the analysis for the non-identifiable model has been recently pointed out [1], [4]. In some models, such as mixture models, the maximum likelihood estimator often diverges. Dacunha-Castelle and Gassiat proposed that the asymptotic behavior of the log likelihood ratio of the maximum likelihood method could be analyzed based on the theory of empirical processes by choosing a locally conic parameterization [2]. Moreover, Hagiwara has shown that the maximum likelihood method makes the generalization error very large, and training error very small [3]. It is well known by many experiments that the Bayesian estimation is more useful than the maximum likelihood method[5].

Recently, we have proven the relation between the Bayesian estimation and the singularities in the parameter space based on algebraic geometry. This relation allows us to analyze non-identifiable models. It reveals that the stochastic complexity depends on the zeta function of the Kullback information from the true distribution to the learning model and of an a priori probability distribution. Using this algebraic geometrical method, we have revealed properties of some models such as multi-layered perceptrons, mixture models and Boltzmann machines.

In this paper, we discuss the application of this method to a Bayesian network. We assume that all latent nodes are directly connected to observable nodes, and there are no connections between all latent nodes. In other words, any hidden node is independent of any other. We obtain the upper bound of the stochastic complexity, which is equal to the minus type II likelihood or the free energy.

## 2 Bayesian Learning and Algebraic Geometry

In this section, we introduce the relation among Bayesian learning, stochastic complexity, and algebraic geometry. Then, we summarize some properties of the stochastic complexity.

### 2.1 Bayesian Learning and Stochastic Complexity

Let $X^n = (X_1, X_2, \cdots, X_n)$ be a set of training samples that are independent and identical. The number of training samples is $n$. These and the testing samples are taken from the true probability distribution $q(x)$. The a priori probability distribution $\varphi(w)$ is given on the set of parameters $W$. Then, the a posteriori probability distribution is defined by

$$p(w|X^n) = \frac{1}{Z_0(X^n)} \varphi(w) \prod_{i=1}^{n} p(X_i|w),$$

where $Z_0(X^n)$ is a normalizing constant. The empirical Kullback information is given by

$$H_n(w) = \frac{1}{n} \sum_{i=1}^{n} \log \frac{q(X_i)}{p(X_i|w)}.$$

Then, $p(w|X^n)$ is rewritten as

$$p(w|X^n) = \frac{1}{Z(X^n)} \exp(-nH_n(w)) \, \varphi(w),$$

where the normalizing constant $Z(X^n)$ is given by

$$Z(X^n) = \int \exp(-nH_n(w))\varphi(w)dw.$$

The stochastic complexity is defined by

$$F(X^n) = -\log Z(X^n).$$

We can select the optimal model and hyperparameters by minimizing $-\log Z_0(X^n)$. This is equivalent to minimizing the stochastic complexity, since

$$-\log Z_0(X^n) = -\log Z(X^n) + S(X^n),$$
$$S(X^n) = -\sum_{i=1}^{n} \log q(X_i),$$

where the empirical entropy $S(X^n)$ is independent of the learners. The average stochastic complexity $F(n)$ is defined by

$$F(n) = -E_{X^n}\left[\log Z(X^n)\right], \quad (1)$$

where $E_{X^n}$ stands for the expectation value over all sets of training samples.

The Bayesian predictive distribution $p(x|X^n)$ is given by

$$p(x|X^n) = \int p(x|w)p(w|X^n)dw.$$

The generalization error $G(n)$ is the average Kullback information from the true distribution to the Bayesian predictive distribution,

$$G(n) = E_{X^n}\left[\int q(x)\log\frac{q(x)}{p(x|X^n)}dx\right].$$

Clarifying the behavior of $G(n)$, when the number of training samples is sufficiently large, is very important. The relation between $G(n)$ and $F(n)$ is

$$G(n) = F(n+1) - F(n). \quad (2)$$

This relation is well known [8] and allows that the generalization error can be calculated from the average stochastic complexity. When $F(n)$ is obtained as

$$F(n) = \lambda \log n,$$

the model's generalization error is given by

$$G(n) = \frac{\lambda}{n}.$$

If a learning machine is an identifiable and regular statistical model, it is proven [7] that asymptotically

$$F(n) = \frac{d}{2}\log n + const.$$

holds, where $d$ is the dimension of the parameter space $W$.

## 2.2 Stochastic Complexity and Algebraic Geometry

We define the Kullback information from the true distribution $q(x)$ to the learner $p(x|w)$ by

$$H(w) = \int q(x)\log\frac{q(x)}{p(x|w)}dx. \quad (3)$$

The asymptotic form of the stochastic complexity strongly relates to the singularities of the parameter set $\{w; H(w) = 0\}$. Note that the important and nontrivial relation was clarified by the algebraic geometrical method [9], [10].

Assume that the Kullback information $H(w)$ is an analytic function of $w$ in the support of the a priori distribution. If the learner is in a redundant state in comparison with the true distribution, the set $\{w \in W; H(w) = 0\}$ includes quite complicated singularities. The algebraic geometry is the only means by which we can analyze the effect of singularities. We need the function $J(z)$ of a complex variable $z$, which is defined by

$$J(z) = \int H(w)^z \varphi(w)dw. \quad (4)$$

This function is called the zeta function of $H(w)$ and the a priori distribution $\varphi(w)$. It is a holomorphic function in the region $Re(z) > 0$, and can be analytically continued to the meromorphic function on the entire complex plane. Its poles are all real, negative and rational numbers. This continuation is ensured by the existence of the b-function.

Let $0 > -\lambda_1 > -\lambda_2 > \cdots$ be the sequence of poles of the zeta function ordered from the origin to minus infinity, and $m_1, m_2, \cdots$ be the respective orders of the poles. That $\mathcal{F}(n)$ defined by

$$\mathcal{F}(n) = -\log\int \exp(-nH(w))\varphi(w)dw$$

is the upper bound of $F(n)$ has been proven. This can be rewritten as

$$\mathcal{F}(n) = \lambda_1 \log n - (m_1 - 1)\log\log n + const. \quad (5)$$

for $n \to \infty$. The coefficient of the leading term in $\mathcal{F}(n)$ is $\lambda_1$, the absolute value of the largest pole. In fact, we can calculate $\lambda_1$ and $m_1$ by using the resolution of singularities in algebraic geometry [9]. However, finding the complete resolution map is generally difficult [11]. We can alternatively find a partial resolution of singularities. This gives us the pole $-\mu$ of zeta function $J(z)$. Then, we obtain the upper bound of the stochastic complexity, since $\mu$ is the upper bound of $\lambda_1$. According to this formula, $\lambda_1$ can be found in some models such as multi-layer neural networks [10] and mixture models [12]. In this paper, we evaluate Bayesian networks, and prove the upper bound of $\lambda$ by finding a pole of the zeta function.

## 2.3 Basic Properties of Stochastic Complexity

Let us summarize some basic properties of stochastic complexity.

First, define a function $\mathcal{F}(S,\psi)$ by

$$\mathcal{F}(S,\psi) = -\log\int \exp(-nS(w))\psi(w)dw,$$

where $S$ is a function of $w$ and $\psi$ is a non-negative function of $w$. This is well defined even if $\psi(w)$ is not a probability density function.

(Proposition. 1) Using Jensen's inequality, we can show easily that the following inequality holds [8],

$$F(n) \leq \mathcal{F}(H, \varphi), \quad (6)$$

where $H(w)$ is the Kullback information defined by the equation (3).

(Proposition. 2) If the functions $H_1, H_2$ and the positive functions $\varphi_1, \varphi_2$ satisfy

$$H_1(w) \leq H_2(w) \quad (\forall w \in W),$$
$$\varphi_1(w) \geq \varphi_2(w) \quad (\forall w \in W),$$

then the following inequality immediately holds,

$$\mathcal{F}(H_1, \varphi_1) \leq \mathcal{F}(H_2, \varphi_2).$$

This inequality also claims that, if the integrated region in the parameter set is $U \supset V$,

$$-\log \int_U \exp(-nK(w))\psi(w)dw$$
$$\leq -\log \int_V \exp(-nK(w))\psi(w)dw$$

holds. Based on this property, if we consider the restricted parameter set, we obtain the upper bound of the stochastic complexity.

(Proposition. 3) Assume that $w = (w_1, w_2)$ and $H$ and $\varphi$ are separated into two functions of each other,

$$H(w_1, w_2) = H_1(w_1) + H_2(w_2),$$
$$\varphi(w_1, w_2) = \varphi_1(w_1) \varphi_2(w_2).$$

The following equality holds,

$$\mathcal{F}(H, \varphi) = \mathcal{F}(H_1, \varphi_1) + \mathcal{F}(H_2, \varphi_2).$$

Define the zeta functions by

$$J(z) = \int H(w)\varphi(w)dw,$$
$$J_i(z) = \int H_i(w_i)\varphi_i(w_i)dw_i \quad (i = 1, 2).$$

Let $-\mu, -\mu_1, -\mu_2$ be the largest poles of $J, J_1$ and $J_2$. This property claims that

$$\mu = \mu_1 + \mu_2. \quad (7)$$

## 3 Main Results

In this section, we introduce Bayesian networks and state the main theorem that clarifies the upper bounds of the stochastic complexities.

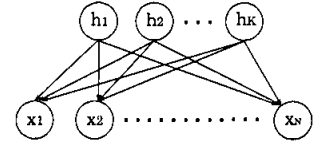

Figure 2: The Bayesian network.

### 3.1 Bayesian networks

Let $x$ be the observed node, and $h = \{h_k\}_{k=1}^{K}$ be hidden nodes. Let us assume each hidden node $h_k$ has $T_k$ states, and describe that $h_k \in \{1, 2, \cdots, T_k\}$. Then, the learning model is defined by

$$p(x|w) = \sum_{i_1=1}^{T_1} \cdots \sum_{i_K=1}^{T_K} a_{1i_1} \cdots a_{Ki_K} F(x|b_{i_1 i_2 \cdots i_K}), \quad (8)$$

and the parameter $w$ is given by

$$w = \{a, b\},$$
$$a = \{a_{k i_k}\} \quad (1 \leq k \leq K, 2 \leq i_k \leq T_k),$$
$$b = \{b_{i_1 i_2 \cdots i_K, j}\} \quad (1 \leq j \leq M).$$

Then,

$$a_{k1} = 1 - \sum_{i=2}^{T_k} a_{ki} \quad (1 \leq k \leq K). \quad (9)$$

The dimension of $w$ is

$$\sum_{k=1}^{K}(T_k - 1) + M \prod_{k=1}^{K} T_k. \quad (10)$$

We show the upper bounds of the stochastic complexities of the model represented by equations (8) and (9).

(Remark. 1) If $F(x|b_{i_1 \cdots i_K})$ is given by

$$F(x|b_{i_1 i_2 \cdots i_K}) = \prod_{j=1}^{M} \prod_{l=1}^{Y_j} (b_{i_1 i_2 \cdots i_K, jl})^{\delta(x_j - l)}, \quad (11)$$

$$\delta(n) = \begin{cases} 1 & (\text{if } n = 0) \\ 0 & (\text{otherwise}) \end{cases},$$

the model coincides with the Bayesian network that has observable nodes $x = \{x_j\}_{j=1}^{N}$. We assume that each node $x_j$ has $Y_j$ states and describe $x_j \in \{1, 2, \cdots, Y_j\}$ (Figure 2). Then,

$$b_{i_1 i_2 \cdots i_K, j1} = 1 - \sum_{l=2}^{Y_j} b_{i_1 i_2 \cdots i_K, jl} \quad (1 \leq j \leq M). \quad (12)$$

The dimension of the parameter in $F(x|b_{i_1 \cdots i_K})$ is

$$M = \sum_{j=1}^{N}(Y_j - 1). \quad (13)$$

## 3.2 Main Results

We assume the following two conditions, (A1) and (A2).

(A1) First, assume that the learning model attains the true distribution. The true distribution has $H$ hidden nodes, and each hidden node $h_k$ has $S_k$ states, where $S_k \leq T_k$. In other words, the true parameters $w^*$ exist such that

$$q(x) = p(x|w^*),$$
$$p(x|w^*) = \sum_{i_1=1}^{S_1} \cdots \sum_{i_H=1}^{S_H} a^*_{1i_1} \cdots a^*_{Hi_H} F(x|b^*_{i_1 \cdots i_H}). \quad (14)$$

Thus,

$$\begin{aligned} w^* &= \{a^*, b^*\}, \\ a^* &= \{a^*_{ki_k}\} \quad (1 \leq k \leq H, 2 \leq i_k \leq S_k), \\ b^* &= \{b^*_{i_1 i_2 \cdots i_H, j}\} \quad (1 \leq j \leq M), \end{aligned}$$

and

$$a^*_{k1} = 1 - \sum_{i=1}^{S_k} a^*_{ki} \quad (1 \leq k \leq H). \quad (15)$$

(Remark. 2) In Bayesian networks,

$$F(x|b^*_{i_1 i_2 \cdots i_H}) = \prod_{j=1}^{M} \prod_{l=1}^{Y_j} (b^*_{i_1 i_2 \cdots i_H, jl})^{\delta(x_j - l)}, \quad (16)$$

$$b^*_{i_1 i_2 \cdots i_H, j1} = 1 - \sum_{l=2}^{Y_j} b^*_{i_1 \cdots i_H, jl} \quad (1 \leq j \leq M). (17)$$

(A2) Second, assume that the a priori probability distribution is positive on the true parameter. For a constant $\epsilon > 0$, let us define the subset of parameter $W(\epsilon) \subset W$ by

$$\begin{aligned} W(\epsilon) = \{&\{a, b\} \in W; \\ &|a_{ki_k} - a^*_{ki_k}| \leq \epsilon \ (1 \leq k \leq H, 2 \leq i_k \leq S_k), \\ &|a_{ki_k}| \leq \epsilon \ \text{(otherwise)}, \\ &|b_{i_1 i_2 \cdots i_H 11 \cdots 1, j} - b^*_{i_1 i_2 \cdots i_H, j}| \leq \epsilon \\ &\quad (1 \leq i_m \leq S_m, 1 \leq m \leq H, 1 \leq j \leq M), \\ &|b_{i_1 i_2 \cdots i_H \cdots i_K, j} - b^*_{11 \cdots 1, j}| \leq \epsilon \ \text{(otherwise)}\}. \end{aligned}$$

Suppose that there is a constant $\epsilon > 0$ such that

$$\inf_{W(\epsilon)} \varphi(w) > 0,$$

where '$\inf_{W(\epsilon)}$' denotes the infimum value of $\varphi(w)$ in $w \in W(\epsilon)$.

**Theorem 1** *Assume the conditions, (A1), and (A2). If the learning machine is given by equations (8) and (9), and the true distribution is given by equations (14) and (15), then for arbitrary natural number $n$, the stochastic complexity satisfies the inequality*

$$F(n) \leq C + \mu \log n,$$

$$\mu = \frac{1}{2} M \prod_{k=1}^{H} S_k - \frac{1}{2} \sum_{k=1}^{H} S_k + \frac{1}{2} H + \sum_{k=1}^{K} T_k - K,$$

*where $C$ is a constant independent of $n$.*

## 4 Proof of Main Theorem

Let us define the Kullback informations by

$$D(i_1 i_2 \cdots i_H || i_1 i_2 \cdots i_H, i_{H+1} \cdots i_K)$$
$$= \int dx F(x|b^*_{i_1 \cdots i_H}) \log \frac{F(x|b^*_{i_1 \cdots i_H})}{F(x|b_{i_1 \cdots i_H i_{H+1} \cdots i_K})}.$$

(Remark. 3) If the model is a Bayesian network, we replace $\int dx$ with $\sum_{x_1=1}^{Y_1} \sum_{x_2=1}^{Y_2} \cdots \sum_{x_M=1}^{Y_M}$. The following proof is correctly derived independently of this replacing.

The Kullback information (3) is rewritten as

$$H(w) = \int dx \left[ \left\{ \prod_{k=1}^{H} \sum_{i_k=1}^{S_k} \right\} a^*_{1i_1} \cdots a^*_{Hi_H} F(x|b^*_{i_1 i_2 \cdots i_H}) \right]$$
$$\times \log \frac{\left\{ \prod_{k=1}^{H} \sum_{i_k=1}^{S_k} \right\} a^*_{1i_1} \cdots a^*_{Hi_H} F(x|b^*_{i_1 \cdots i_H})}{\left\{ \prod_{k=1}^{K} \sum_{i_k=1}^{T_k} \right\} a_{1i_1} \cdots a_{Ki_K} F(x|b_{i_1 \cdots i_K})},$$

where

$$\left\{ \prod_{k=1}^{K} \sum_{i_k=1}^{T_k} \right\} = \sum_{i_1=1}^{T_1} \sum_{i_2=1}^{T_2} \cdots \sum_{i_K=1}^{T_K}.$$

Let us divide the parameter $w$ into $w = \{w_1, w_2\}$, where

$$\begin{aligned} w_1 &= \{a_{ki_k}; 1 \leq k \leq H, 2 \leq i_k \leq S_k, \\ &\quad b_{i_1 i_2 \cdots i_K, j}; 1 \leq i_k \leq S_k, 1 \leq k \leq H, \\ &\quad i_{H+1} = i_{H+2} = \cdots = i_K = 1, \\ &\quad 1 \leq j \leq M\}, \\ w_2 &= \{a_{ki}, b_{i_1 i_2 \cdots i_K, j}; \text{otherwise}\}. \end{aligned}$$

Define two functions,

$$H_1(w_1) = \left\{ \prod_{k=1}^{H} \sum_{i_k=1}^{S_k} \right\} \left[ a^*_{1i_1} \cdots a^*_{Hi_H} \log \frac{a^*_{1i_1} \cdots a^*_{Hi_H}}{\gamma_{1i_1} \cdots \gamma_{Hi_H}} \right]$$
$$+ \left\{ \prod_{k=1}^{H} \sum_{i_k=1}^{S_k} \right\} a^*_{1i_1} \cdots a^*_{Hi_H}$$
$$\times D(i_1 \cdots i_H || i_1 \cdots i_H, 11 \cdots 1),$$

$$H_2(w_2) = \sum_{k=1}^{H} \sum_{i_k=S_k+1}^{T_k} c_{ki_k} a_{ki_k} + \sum_{k=H+1}^{K} \sum_{i_k=1}^{T_k} c_{ki_k} a_{ki_k}$$

$$+ c_0 \prod_{k=1}^{H} a_{k1}^* \left\{ \prod_{k=1}^{K} \sum_{i_k=1}^{T_k} \right\} \chi(i_1, \cdots, i_K)$$
$$\times D(1 \cdots 1 || i_1 \cdots i_H, i_{H+1} \cdots i_K),$$

where $\{c_{k i_k}\}$ and $c_0$ are positive constants and

$$\gamma_{k i_k} = \begin{cases} a_{k i_k} & (i_k \neq 1) \\ 1 - \sum_{i_k'=2}^{S_k} a_{k i_k'} & (i_k = 1) \end{cases},$$

$$\chi(i_1, \cdots, i_K) = \begin{cases} 0 & (a_{k i_k} \in W_1; 1 \leq k \leq H, \\ & i_{H+1} = \cdots = i_K = 1) \\ a_{\bar{k} i_{\bar{k}}} & (\bar{k} = \min_k \{k; a_{k i_k} \in W_2\}) \end{cases}.$$

Let us prove the following lemma,

**Lemma 1** *For arbitrary $w \subset W(\epsilon)$,*

$$H(w) \leq H_1(w_1) + H_2(w_2).$$

(Proof of Lemma 1)

In this proof, we use the notations,

$$\sum_1 \equiv \left\{ \prod_{k=1}^{H} \sum_{i_k=1}^{S_k} \right\} \left\{ \prod_{k=H+1}^{K} \sum_{i_k=1}^{1} \right\} - \left\{ \prod_{k=1}^{K} \sum_{i_k=1}^{1} \right\},$$

$$\sum_2 \equiv \left\{ \prod_{k=1}^{K} \sum_{i_k=1}^{T_k} \right\} - \left\{ \prod_{k=1}^{H} \sum_{i_k=1}^{S_k} \right\} \left\{ \prod_{k=H+1}^{K} \sum_{i_k=1}^{1} \right\}$$
$$+ \left\{ \prod_{k=1}^{K} \sum_{i_k=1}^{1} \right\}.$$

In general, the following log-sum inequality holds: For arbitrary sequences of positive numbers $\{d_k, k = 1, 2, \cdots, I\}$ and $\{e_k, k = 1, 2, \cdots, I\}$,

$$\left\{ \sum_{k=1}^{I} d_k \right\} \log \frac{\left\{ \sum_{k=1}^{I} d_k \right\}}{\left\{ \sum_{k=1}^{I} e_k \right\}} \leq \sum_{k=1}^{I} \left\{ d_k \log \frac{d_k}{e_k} \right\}.$$

Thus, for arbitrary sequences of positive numbers $\{d_k, k = 1, 2, \cdots, I\}$ and $\{e_k, k = 1, 2, \cdots, I'\}$, where $I < I'$, it follows that

$$\left\{ \sum_{k=1}^{I} d_k \right\} \log \frac{\left\{ \sum_{k=1}^{I} d_k \right\}}{\left\{ \sum_{k=1}^{I'} e_k \right\}}$$
$$\leq \left\{ \sum_{k=1}^{I} - \sum_{k=1}^{1} \right\} \left\{ d_k \log \frac{d_k}{e_k} \right\}$$
$$+ d_1 \log \frac{d_1}{\left\{ \sum_{k=1}^{I'} - \sum_{k=1}^{I} + \sum_{k=1}^{1} \right\} e_k}.$$

Using this inequality, we obtain

$$H(w) \leq \int dx \left[ \sum_1 a_{1 i_1}^* \cdots a_{H i_H}^* F(x | b_{i_1 \cdots i_H}^*) \right.$$
$$\times \log \frac{a_{1 i_1}^* \cdots a_{H i_H}^* F(x | b_{i_1 \cdots i_H}^*)}{a_{1 i_1} \cdots a_{K1} F(x | b_{i_1 \cdots i_K})}$$
$$+ a_{11}^* a_{21}^* \cdots a_{H1}^* F(x | b_{11 \cdots 1}^*)$$
$$\left. \times \log \frac{a_{11}^* a_{21}^* \cdots a_{H1}^* F(x | b_{11 \cdots 1}^*)}{Z_1(x)} \right],$$

where

$$Z_1(x) = \sum_2 a_{1 i_1} a_{2 i_2} \cdots a_{K i_K} F(x | b_{i_1 i_2 \cdots i_K}).$$

Let us define two functions,

$$R_1(w) = \int dx \left[ \sum_1 a_{1 i_1}^* \cdots a_{H i_H}^* F(x | b_{i_1 i_2 \cdots i_H}^*) \right.$$
$$\left. \times \log \frac{a_{1 i_1}^* \cdots a_{H i_H}^* F(x | b_{i_1 \cdots i_H}^*)}{a_{1 i_1} \cdots a_{K1} F(x | b_{i_1 \cdots i_K})} \right],$$

$$R_2(w) = \int dx \, a_{11}^* a_{21}^* \cdots a_{H1}^* F(x | b_{11 \cdots 1}^*)$$
$$\times \log \frac{a_{11}^* a_{21}^* \cdots a_{H1}^* F(x | b_{11 \cdots 1}^*)}{Z_1(x)}.$$

Then,
$$H(w) \leq R_1(w) + R_2(w). \quad (18)$$

Let us use the following notations,

$$\rho_{i_1 i_2 \cdots i_K} = \frac{a_{i_1} a_{i_2} \cdots a_{i_K}}{\sigma},$$
$$\sigma = \sum_2 a_{1 i_1} a_{2 i_2} \cdots a_{K i_K}.$$

Then, $\rho_{i_1 i_2 \cdots i_K}$ is a probability distribution. We can rewrite $R_2(w)$ as

$$R_2(w) = \sum_x a_{11}^* a_{21}^* \cdots a_{H1}^* F(x | b_{11 \cdots 1}^*)$$
$$\times \left\{ \log \frac{a_{11}^* a_{21}^* \cdots a_{H1}^*}{\sigma} + \log \frac{F(x | b_{11 \cdots 1}^*)}{Z_2(x)} \right\},$$

where

$$Z_2(x) = \sum_2 \rho_{i_1 i_2 \cdots i_K} F(x | b_{i_1 i_2 \cdots i_K}).$$

Applying Jensen's inequality to $R_2(w)$, we obtain

$$R_2(w) \leq \sum_2 \rho_{i_1 \cdots i_K} D(11 \cdots 1 || i_1 \cdots i_H, i_{H+1} \cdots i_K)$$
$$+ a_{11}^* a_{21}^* \cdots a_{H1}^* \log \frac{a_{11}^* a_{21}^* \cdots a_{H1}^*}{\sigma}. \quad (19)$$

In the region $W(\epsilon)$, a positive constant $c_1$ exists such that

$$\rho_{i_1 i_2 \cdots i_K} \leq \frac{a_{k i_k}}{\sigma} \quad (1 \leq \forall k \leq K)$$
$$\leq \frac{a_{k i_k}}{c_1}. \quad (20)$$

We can easily obtain

$$a_{k1}^* \log \frac{a_{k1}^*}{a_{k1}} \leq \left(1 - \sum_{i_k=2}^{S_k} a_{ki_k}^*\right) \log \frac{1 - \sum_{i_k=2}^{S_k} a_{ki_k}^*}{1 - \sum_{i_k=2}^{S_k} a_{ki_k}} + c_{2k} \sum_{i_k=S_k+1}^{T_k} a_{ki_k}, \quad (21)$$

for $w \in W(\epsilon)$ and $k \leq H$, where $c_{2k}$ is a positive constant, and

$$\log \frac{1}{a_{k1}} \leq c_{3k} \sum_{i_k=S_k+1}^{T_k} a_{ki_k}, \quad (22)$$

for $w \in W(\epsilon)$ and $k \geq H+1$, where $c_{3k}$ is a positive constant. In the inequality (19), each $\rho_{i_1 \cdots i_K}$ has $a_{ki_k} \in W_2$ as the factor. Using $0 \leq \rho_{i_1 i_2 \cdots i_K} \leq 1$, (20), (21) and (22), we can obtain

$$R_1(w) \leq \sum_{k=1}^{H} \sum_{i_k=S_k+1}^{T_k} c''_{ki_k} a_{ki_k} + \sum_{k=H+1}^{K} \sum_{i_k=2}^{T_k} c''_{ki_k} a_{ki_k}$$
$$+ \sum_1 a_{1i_1}^* a_{2i_2}^* \cdots a_{Hi_H}^* \log \frac{a_{1i_1}^* a_{2i_2}^* \cdots a_{Hi_H}^*}{\gamma_{1i_1} \gamma_{2i_2} \cdots \gamma_{Hi_H}}$$
$$+ \sum_1 a_{1i_1}^* \cdots a_{Hi_H}^* D(i_1 \cdots i_H \| i_1 \cdots i_H, 1 \cdots 1),$$

$$R_2(w) \leq a_{11}^* a_{21}^* \cdots a_{H1}^* \log \frac{a_{11}^* a_{21}^* \cdots a_{H1}^*}{\gamma_{11} \gamma_{22} \cdots \gamma_{H1}}$$
$$+ \sum_{k=1}^{H} \sum_{i_k=S_k+1}^{T_k} c'_{ki_k} a_{ki_k} + \sum_{k=H+1}^{K} \sum_{i_k=2}^{T_k} c'_{ki_k} a_{ki_k}$$
$$+ \frac{1}{c_1} \left\{ \prod_{k=1}^{K} \sum_{i_k=1}^{T_k} \right\} \chi(i_1, i_2, \cdots, i_K)$$
$$\times D(11 \cdots 1 \| i_1 i_2 \cdots i_H, i_{H+1} \cdots i_K)$$
$$+ D(11 \cdots 1 \| 11 \cdots 1, 11 \cdots 1),$$

where $\{c'_{ki_k}\}$ and $\{c''_{ki_k}\}$ are positive constants. By combining the above inequalities with (18), we obtain Lemma 1. (End of Proof)

Let us define two sets of the parameters

$$W_1 = \{w_1; |a_{ki_k} - a_{ki_k}^*| \leq \epsilon$$
$$(1 \leq k \leq H, 2 \leq i_k \leq S_k),$$
$$|b_{i_1 i_2 \cdots i_H 11 \cdots 1, j} - b_{i_1 i_2 \cdots i_H, j}^*| \leq \epsilon$$
$$(1 \leq k \leq H, 1 \leq i_k \leq S_k, 1 \leq j \leq M)\},$$
$$W_2 = \{w_2; |a_{ki_k}| \leq \epsilon, |b_{i_1 i_2 \cdots i_K, j} - b_{11 \cdots 1, j}^*| \leq \epsilon$$
$$(\text{otherwise})\}.$$

Here, $w_1 \in W_1$ and $w_2 \in W_2$ are free variables. Also, define the partial stochastic complexities,

$$F_i(n) = -\log \int_{W_i'} \exp(-nH_i(w_i)) dw_i \quad (i = 1, 2),$$

where the integrated region $W_1$ and $W_2$ are taken such that $W_1' \subset W_1$ and $W_2' \subset W_2$, and that

$$W_1' \times W_2' \subset \text{supp}\varphi(w),$$

where $\varphi(w)$ is the support of the a priori distribution. From the assumption (A2),

$$\eta \equiv \inf_{w \in W_1 \times W_2} \varphi(w) > 0.$$

The stochastic complexity is bounded by

$$F(n) \leq -\log \eta - \sum_{i=1}^{2} \log \int_{W_i} \exp(-nH_i(w_i)) dw_i.$$

Thus,

$$F(n) \leq F_1(n) + F_2(n) + const.$$

In order to prove Theorem 1, it is sufficient to bound each $F_i(n)$ $(i = 1, 2)$. To bound $F_1(n)$ is easy, because it can be bounded by the stochastic complexity of identifiable models. Thus, we obtain Lemma 2.

**Lemma 2** *A partial stochastic complexity satisfies the inequality,*

$$F_1(n) \leq \frac{1}{2} \left\{ M \prod_{k=1}^{H} S_k + \sum_{k=1}^{H} (S_k - 1) \right\} \log n + C_1,$$

*where $C_1$ is a constant independent of $n$.*

However, because the set $\{w_2; H_2(w_2) = 0\}$ includes singularities, we apply the algebraic geometrical method to $F_2(n)$.

**Lemma 3** *The second partial stochastic complexity satisfies the inequality,*

$$F_2(n) \leq \left\{ \sum_{k=1}^{H} (T_k - S_k) + \sum_{k=H+1}^{K} (T_k - 1) \right\} \log n + C_2,$$

*where $C_2$ is a constant independent of $n$.*

(Proof of Lemma 3)

In order to clarify the asymptotic expansion of $F_2(n)$, we consider the zeta function,

$$J(z) = \int_{W_2} H_2(w_2)^z dw_2.$$

Based on the algebraic geometrical method, we need to show that this zeta function has a pole,

$$z = -\left\{ \sum_{k=1}^{H} (T_k - S_k) + \sum_{k=H+1}^{K} (T_k - 1) \right\}.$$

According to the definition of $\chi(i_1, i_2, \cdots, i_K)$, all the terms of $H_2(w_2)$ have $a_{k i_k}$ as the factor. Now, we define a variable $w_3$ and a mapping

$$g: w_3 = (\omega, \{\omega_{k i_k}\}, \{b_{i_1 i_2 \cdots i_K, j}\}) \mapsto w_2$$

by

$$\begin{aligned}
\omega &= a_{K T_K}, \\
\omega \omega_{k i_k} &= a_{k i_k} \quad (1 \leq k \leq H, S_k + 1 \leq i_k \leq T_k), \\
\omega \omega_{k i_k} &= a_{k i_k} \quad (H+1 \leq k \leq K-1, 2 \leq i_k \leq T_k), \\
\omega \omega_{K i_K} &= a_{K i_K} \quad (2 \leq i_K \leq T_K - 1).
\end{aligned}$$

This mapping is called a blow-up in algebraic geometry. The function $H(g(w_3))$ divided $\omega$ is a constant function of $\omega$,

$$H_3(\{\omega_{k i_k}\}, \{b_{i_1 \cdots i_K, j}\}) \equiv H_2(w_2)/\omega.$$

The Jacobian $|g'(w_3)|$ of the mapping $g$ is

$$\begin{aligned}
|g'(w_3)| &= \omega^{\bar{d}_2}, \\
\bar{d}_2 &\equiv \sum_{k=1}^{H}(T_k - S_k) + \sum_{k=H+1}^{K}(T_k - 1) - 1.
\end{aligned}$$

Thus, we can integrate the variable $\omega$,

$$\begin{aligned}
J(z) &= \int_0^\epsilon \omega^{z+\bar{d}_2} \hat{J}(z) d\omega \\
&= \frac{\epsilon^{z+\bar{d}_2}}{z+\bar{d}_2+1} \hat{J}(z), \\
\hat{J}(z) &= \int H_3(\{\omega_{k i_k}\}, \{b_{i_1 \cdots i_K, j}\})^z \prod d\omega_{k i_k} \prod db_{i_1 \cdots i_K, j}.
\end{aligned}$$

If $z$ is real and larger than the largest pole of $\hat{J}(z)$, the function $\hat{J}(z)$ is not equal to zero. Thus the largest pole of $J(z)$ is not smaller than $z = -(\bar{d}_2 + 1)$, which completes the proof of Lemma 3. (End of Proof)

Combining Lemma 1-3 with the properties of the stochastic complexity (Proposition. 2, 3), we obtain Theorem 1. (End of Proof)

## 5　Discussion & Conclusion

Let us apply Theorem 1 to Bayesian networks. As we mentioned at **(Remark.3)**, the proof of Theorem 1 is correctly derived even if $x$ is a discrete random variable. Therefore, if a Bayesian network is given by the equations (8), (9), (11) and (12), and the true distribution is given by the equations (14), (15), (16) and (17), then the stochastic complexity has the coefficient,

$$\mu = \frac{1}{2}\sum_{j=1}^{N}(Y_j - 1)\prod_{k=1}^{H}S_k - \frac{1}{2}\sum_{k=1}^{H}S_k + \frac{1}{2}H + \sum_{k=1}^{K}T_k - K.$$

This result requires us to improve BIC. From our result, the average stochastic complexity of the Bayesian network is far smaller than $(d/2)\log n$, where $d$ is the dimension of the parameter space. This means that we cannot select the optimal sized model when we apply Schwartz's BIC to this model. The same results were clarified where the learner has one binary hidden node and binary observable nodes [6]. We trust that our result provides the mathematical foundation for an improved information criterion.

### Acknowledgment

This work was partially supported by the Ministry of Education, Science, and Culture in Japan, Grant-in-aid for scientific research 15500130 and for JSPS Fellows 3544, 2003.